\documentclass[letterpaper]{article}

\ifdefined\aaaianonymous
    \usepackage[submission]{aaai2026}
\else
    \usepackage{aaai2026}
\fi

\usepackage{times}
\usepackage{helvet}

\usepackage[hyphens]{url}
\usepackage{graphicx}
\usepackage{natbib}
\usepackage{caption}
\usepackage{amsmath,amssymb}
\usepackage{booktabs}
\usepackage{multirow}
\usepackage{subcaption}
\usepackage{xcolor}
\usepackage{colortbl}
\usepackage{pifont}

\newcommand{\method}{Poly-OPD}
\newcommand{\scorehl}[1]{\begingroup\setlength{\fboxsep}{0pt}\colorbox{blue!10}{#1}\endgroup}

\newcommand{\teasercaption}{Motivation and outcome of heterogeneous teacher
distillation. Left: FLUX.1-dev produces visually appealing images but can miss
prompt details such as spatial relations, colors, or object counts, whereas
Z-Image better follows these compositional constraints but often has weaker
visual appeal. Right: the radar chart compares FLUX.1-dev, Z-Image, the
SD3.5-Medium base student, and \method{}, showing that the distilled student
combines stronger preference and composition capabilities within one selectable
model.}

\title{\method{}: Heterogeneous Multi-Teacher On-Policy Distillation for Capability-Selectable Flow Models}

\author{
    Siming Fu\textsuperscript{\rm 1}\textsuperscript{*},
    Haojun Xu\textsuperscript{\rm 1}\textsuperscript{*},
    Ruizhe He\textsuperscript{\rm 1}\textsuperscript{*},
    Zheming Fu\textsuperscript{\rm 1}\textsuperscript{*},
    Hualiang Wang\textsuperscript{\rm 2},
    Jie Huang\textsuperscript{\rm 1},\\
    Xiaoxiao Ma\textsuperscript{\rm 1},
    Mingchen Zhong\textsuperscript{\rm 1},
    Weihu Huang\textsuperscript{\rm 2},
    Xiaoxuan He\textsuperscript{\rm 2},
    Linjiang Huang\textsuperscript{\rm 3},
    Si Liu\textsuperscript{\rm 3}\textsuperscript{\dag}
}
\affiliations{
    \textsuperscript{\rm 1}Joy Future Academy\quad
    \textsuperscript{\rm 2}Zhejiang University\quad
    \textsuperscript{\rm 3}Beihang University\\
    \textsuperscript{*}Equal contribution.\quad
    \textsuperscript{\dag}Corresponding author.\\
    fusiming.chosen@jd.com\quad   liusi@buaa.edu.cn\\
}

\makeatletter
\def\@maketitle{%
  \def\theauthors{\if T\showauthors@on\@author\else Anonymous submission\fi}%
  \newcounter{eqfn}\setcounter{eqfn}{0}%
  \vbox{%
    \let\footnote\thanks\relax%
    \setcounter{footnote}{0}%
    \def\equalcontrib{%
      \ifnum\value{eqfn}=0%
        \footnote{These authors contributed equally.}%
        \setcounter{eqfn}{\value{footnote}}%
      \else%
        \footnotemark[\value{eqfn}]%
      \fi%
    }%
    \hsize\textwidth%
    \linewidth\hsize%
    \vskip 0.625in minus 0.125in%
    \centering%
    {\LARGE\bf \@title \par}%
    \vskip 0.1in plus 0.5fil minus 0.05in%
    {\Large{\textbf{\theauthors\ifhmode\\\fi}}}%
    \vskip .2em plus 0.25fil%
    {\normalsize \affiliations_\ifhmode\\\fi}%
    \vskip 0.7em%
    \includegraphics[width=0.98\textwidth]{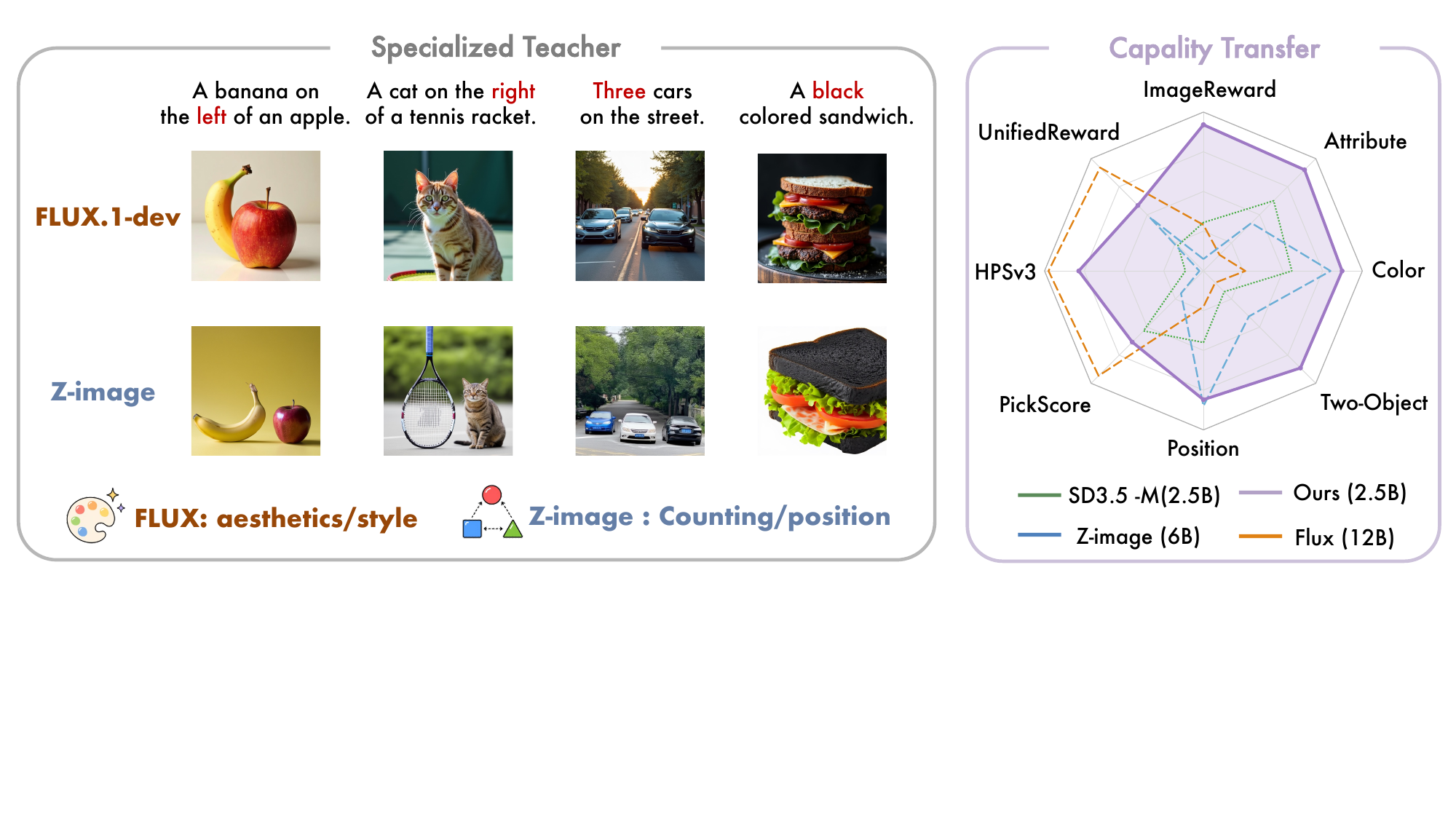}\par%
    \vspace{-2.4cm}%
    \refstepcounter{figure}\label{fig:teaser}%
    \noindent\makebox[\textwidth][c]{%
      \begin{minipage}{0.96\textwidth}%
      \small\textbf{Figure~\thefigure:} \teasercaption%
      \end{minipage}%
    }\par%
    \vskip 1em plus 2fil%
  }%
}%
\makeatother

\nocopyright
\begin{document}
\maketitle

\begin{abstract}
Leading open text-to-image models often carry complementary strengths:
one may lead on preference-aligned aesthetics while another follows
compositional instructions more faithfully. However, differences in their autoencoders and noise schedules make it difficult to transfer these strengths across models. In this paper, we present \textbf{Poly-OPD}, a framework that can consolidate complementary strengths of heterogeneous teachers into a single compact flow-matching student. 
To bridge the incompatible latent spaces of different teachers, Poly-OPD performs \emph{on-policy} distillation through a pixel bridge. Each student-generated image is re-encoded by a selected teacher's encoder and refined from a noise level matched by magnitude under the teacher's noise schedule. The resulting target is further matched to the student in frozen DINOv2 space, enabling supervision across incompatible latent spaces.
To retain complementary capabilities without cross-teacher interference, Poly-OPD uses a gradient compatibility diagnostic to organize its adapters: attention LoRA modules are shared across teachers, whereas feed-forward adapters remain teacher-specific. During distillation, a gap-aware curriculum devotes more training to compositional categories where the student still falls short of the teacher. As each gap narrows, training shifts toward categories with larger remaining gaps. By distilling FLUX.1-dev and Z-Image into a 2.5B SD3.5-Medium student, Poly-OPD improves GenEval from 67.3 to 73.3, surpassing both larger teachers, and raises DrawBench HPSv3 from 9.34 to 11.35, consolidating both strengths within a switchable model.
\end{abstract}

\section{Introduction}
\label{sec:intro}

Text-to-image generation is increasingly served by specialists. Some
models are aligned to human preference and excel at visual quality and
aesthetics~\cite{xu2023imagereward, wu2023human}; others are optimized for compositional instruction following,
covering subskills such as counting, attribute binding, and spatial
relations~\cite{hu2024ella,ghosh2023geneval}; the two strengths rarely coexist in one
checkpoint. 
Worse, these specialists are built on different backbones~\cite{esser2024sd3,flux2024,cai2025z}: they use
different autoencoders, denoisers, and noise schedules, so their latent
spaces are mutually unintelligible. An application that needs both
capabilities today must deploy multiple large models and route between
them, paying the memory, latency, and maintenance cost of each.

We study the problem of \textbf{multi-capability consolidation}: given a
set of frozen, architecturally heterogeneous teachers, each expert in one
capability, distill them into a single flow-matching student such that
(i) each capability matches its teacher, (ii) capabilities do not degrade
one another, and (iii) the active capability is selectable at inference
at negligible cost. This setting breaks the central assumption of
existing distillation. Score, velocity, and trajectory matching all
supervise the student in the teacher's coordinate
system~\cite{salimans2022progressive,luo2023lcd,zhou2024score,zhao2026mean}, 
and therefore require a shared autoencoder and noise schedule; a
heterogeneous teacher offers neither, so its latents and trajectories are
simply not valid targets. The common fallback of training on
teacher-generated images sidesteps the coordinate problem but is
\emph{off-policy}: the student is supervised on states derived from
teacher outputs, yet at inference it follows its own trajectory, and the
resulting train--inference mismatch caps how closely it can track the
teacher~\cite{agarwal2024policy}. 
Two further difficulties compound the setting. Capabilities distilled
into one network interfere~\cite{yu2020gradient}: we find preference- and composition-mode
gradients pointing in opposing directions in parts of the network.
Moreover, the compositional capability is itself imbalanced, with some
subskills saturating early while others remain bottlenecks, so uniform
training budget is systematically misallocated.

We propose \textbf{\method{}}, which resolves all three difficulties. Its core is
\textbf{heterogeneous on-policy distillation}: the student's own sample
is decoded to pixels, the only coordinate system heterogeneous models
share, then re-encoded into the active teacher's latent space and refined
by the frozen teacher from a noise level matched by magnitude rather than
timestep index. The refined image is a teacher correction of what the
student actually produced, i.e.\ an \emph{on-policy} target obtained from
a teacher whose internal coordinates the student cannot read, and the
refinement depth interpolates continuously between off-policy
distillation on teacher samples and local correction of student samples.
Supervision is applied in a frozen DINOv2 feature space~\cite{oquab2023dinov2}, whose invariance
to model-specific pixel statistics is what makes comparison across
latent-incompatible models well posed. To make capabilities coexist,
\method{} employs \textbf{capability-selectable adapters} and sets their
sharing--isolation boundary by measurement instead of convention: a
gradient-compatibility diagnostic shows that attention updates transfer
across capabilities while feed-forward updates conflict, so \method{}
shares a single attention LoRA~\cite{hu2022lora} and isolates per-capability FFN adapters.
As a result, one frozen backbone carries all capabilities, and selecting
one costs an adapter swap. Finally, \textbf{gap-aware sampling} allocates
compositional prompts by the remaining teacher--student margin rather
than raw difficulty, so budget flows to categories where the teacher
still has something to teach and anneals away as gaps close. \textbf{Our contributions are as follows:}
\begin{itemize}
    \item We formulate \textbf{\emph{multi-capability consolidation}}: distilling
    architecturally heterogeneous, latent-incompatible teachers into one
    flow-matching student whose active capability is selectable at
    inference, a setting in which existing latent-space distillation is
    inapplicable by construction.
    \item We propose \textbf{\method{}}, which obtains \emph{on-policy}
    supervision from such teachers through a pixel bridge with
    noise-magnitude alignment and semantic-space supervision, separates
    conflicting capabilities with adapters whose sharing--isolation
    boundary is set by a gradient-compatibility measurement, and
    allocates compositional budget through gap-aware sampling driven by
    the remaining teacher--student residual.
    \item Distilling FLUX.1-dev~\cite{flux2024} and Z-Image~\cite{cai2025z} into a 2.5B SD3.5-Medium~\cite{esser2024sd3}
    student, \method{} raises GenEval~\cite{ghosh2023geneval} from 67.3 to 73.3, surpassing both
    larger teachers, and raises HPSv3~\cite{ma2025hpsv3} on
    DrawBench~\cite{saharia2022drawbench} from 9.34 to 11.35, all within
    one backbone where switching capabilities costs an adapter swap. 
\end{itemize}

\section{Related Work}
\label{sec:related}

\paragraph{Diffusion distillation.}
Diffusion and flow-matching models achieve high-quality generation
through iterative denoising or continuous transport from noise to
data~\cite{ho2020ddpm,song2021scorebased,lipman2023flow,
liu2023flow}. Existing distillation methods accelerate generation
through trajectory matching, consistency training, adversarial
objectives, or distribution matching
~\cite{salimans2022progressive,luo2023lcd,
song2024improved,yin2024dmd,yin2024dmd2}.
These methods primarily transfer a single teacher into a compatible
few-step student.

Recent on-policy methods instead supervise states visited by the
current student. DiffusionOPD derives transition-level objectives for
diffusion models, while Flow-OPD uses dense velocity supervision to
integrate task-specialized flow teachers
~\cite{li2026diffusionopd,fang2026flow}.
CollectionLoRA further consolidates multiple customized LoRA teachers
and few-step generation into a shared adapter
~\cite{wu2026collectionlora}.
These approaches assume compatible transition kernels, vector fields,
or model-derived adapters.

\paragraph{Multi-teacher adaptation.}
Multi-teacher knowledge distillation transfers complementary
knowledge from several teachers into a compact student
~\cite{you2017learning,gu2024minillm,ko2025distillm}. Knowledge amalgamation extends this setting
to teachers trained for different tasks or built with heterogeneous
architectures, typically through feature transformation or selective
teacher supervision
~\cite{shen2019amalgamating,jing2021amalgamating}.
Joint adaptation to multiple capabilities can suffer from destructive
interference and uneven learning progress. Multi-task methods address
these issues through gradient normalization, conflict projection, or
dynamic task weighting
~\cite{chen2018gradnorm,yang2026learning,jin2026entropy},
whereas modular approaches isolate task-specific knowledge using
adapters or LoRA experts
~\cite{hu2022lora,feng2024mixture}.

\begin{figure*}[!h]
\centering
\includegraphics[width=0.98\textwidth]{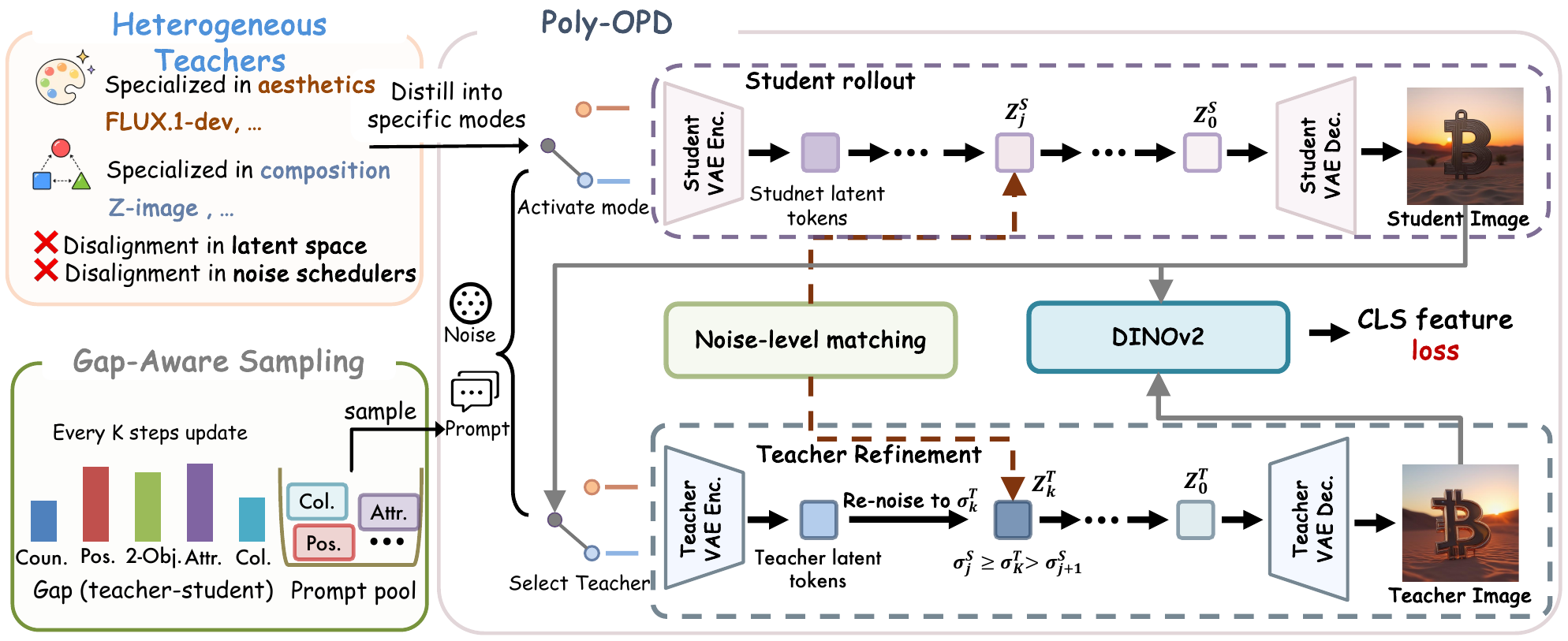}
\caption{Overview of \method{}. The current student sample is bridged
through pixels into the selected teacher latent space, refined by the
teacher from a matched noise level, and used as perceptual supervision in
a common DINOv2 feature space. Capability-selectable adapters isolate
mode-specific FFN updates while sharing attention LoRA, and gap-aware
sampling reallocates compositional prompts toward categories with the
largest teacher--student deficits.}
\label{fig:pipeline}
\end{figure*}

\section{Method}
\label{sec:method}
\textbf{\method{}} consolidates heterogeneous, latent-incompatible teachers into a single multi-capability student by addressing three key challenges. First, differences in autoencoders, denoisers, and noise schedules eliminate any shared latent coordinate system, making direct supervision with teacher latents or trajectories infeasible. We therefore introduce \textbf{\textit{heterogeneous on-policy distillation}}, which bridges the student and teachers through pixel space and applies supervision in a semantic feature space (Sec.~\ref{sec:hetero-opd}). Second, jointly distilling distinct capabilities can cause destructive interference. To mitigate this issue, \textbf{\textit{capability-selectable adapters}} use gradient compatibility to determine which adaptations should be shared and which should remain capability-specific (Sec.~\ref{sec:moe}). Third, uniform prompt sampling wastes training budget on already-mastered subskills. \textbf{\textit{Gap-aware adaptive sampling}} instead reallocates prompts according to the remaining teacher--student gap (Sec.~\ref{sec:gap-aware}).

\subsection{Problem Setup}
\label{sec:problem}

Let $S_\theta$ be a flow-matching student~\cite{lipman2023flow,liu2023flow} and $\mathcal{M}=\{1,\ldots,M\}$
a set of capability modes, where each mode $e$ pairs a frozen teacher
$T_e$ with a prompt source $\mathcal{P}_e$. During training the sampled
supervision source determines the active mode; at inference the
application selects it. A mode may contain fine-grained categories, such
as counting or color attribution within the compositional mode, but these
are not separate teachers. Crucially, teachers are architecturally
unconstrained: they may use different autoencoders, denoisers, and noise
schedules, so their latents and trajectories are not valid coordinate
targets for the student. The task is therefore threefold: (i) obtain
\emph{on-policy} supervision from latent-incompatible teachers, (ii)
expose \emph{selectable} capabilities within one student without mutual
degradation, and (iii) allocate training budget \emph{within} a mode
according to what remains to be learned. Sections~\ref{sec:hetero-opd},
\ref{sec:moe}, and \ref{sec:gap-aware} address these in turn.

\paragraph{Notation.}
$\mathcal{E}_S,\mathcal{G}_S$ are the student VAE encoder and decoder,
and $\mathcal{E}_e,\mathcal{G}_e$ those of $T_e$; $v_\theta$ is the
student velocity field and $\mathrm{sg}[\cdot]$ the stop-gradient. Noise
schedules are indexed from noise to data,
$1=\sigma_0^S>\cdots>\sigma_{N_S}^S=0$ and likewise
$\{\sigma_i^e\}_{i=0}^{N_e}$, so sampling proceeds with increasing index.
Module types are $u\in\{\mathrm{attn},\mathrm{ffn}\}$ and blocks
$l\in\{1,\ldots,L\}$.

\subsection{Heterogeneous On-Policy Distillation}
\label{sec:hetero-opd}

Figure~\ref{fig:pipeline} summarizes the pipeline: a warm start brings
the student near the teacher distribution, after which on-policy
distillation through a pixel bridge closes the remaining
train--inference gap.

\paragraph{Stage 1: warm start.}
For $c\sim\mathcal{P}_e$, the teacher produces
$x_e^\star=\mathrm{Sample}_{T_e}(c)$, which we re-encode into student
coordinates as $z_{\mathrm{clean}}^{S,e}=\mathcal{E}_S(x_e^\star)$; the
teacher acts here purely as a data generator. With
$\epsilon\sim\mathcal{N}(0,I)$ and $\sigma$ from the student schedule, the
flow-matching interpolation~\cite{lipman2023flow,liu2023flow}
$z_\sigma^S=(1-\sigma)z_{\mathrm{clean}}^{S,e}+\sigma\epsilon$ gives
target velocity $\epsilon-z_{\mathrm{clean}}^{S,e}$, and the
mode-conditional objective is
\begin{equation}
    \mathcal{L}_{\mathrm{WS}}^{e}
    =
    \mathbb{E}_{c\sim\mathcal{P}_e,\sigma,\epsilon}\!\left[
    \big\|v_\theta(z_\sigma^S,\sigma,c)-(\epsilon-z_{\mathrm{clean}}^{S,e})\big\|_2^2
    \right].
    \label{eq:warm-start-loss}
\end{equation}

\paragraph{Stage 2: on-policy distillation via a pixel bridge.}
Warm start remains off-policy: it trains on states derived from teacher
outputs, while inference follows the student's own trajectory~\cite{li2026diffusionopd}. Following the on-policy principle of supervising states generated by
the student itself Stage 2
instead queries the active teacher with the student's own sample~\cite{li2026diffusionopd,fang2026flow}. The
student draws $z_0^S\sim\mathcal{N}(0,I)$ and rolls out without
gradients,
$\{z_i^S\}_{i=0}^{N_S}=\mathrm{Euler}_S(z_0^S,0\!\rightarrow\!N_S,c;\theta)$.
Since pixels are the only coordinate system the two models share, we
decode the endpoint and bridge it into the teacher latent space,
\begin{equation}
    x_S=\mathcal{G}_S(z_{N_S}^S),
    \qquad
    \bar{z}_{\mathrm{clean}}^{e}=\mathcal{E}_e(x_S).
    \label{eq:pixel-bridge}
\end{equation}
Sampling a refinement depth
$r\sim\mathcal{U}\{r_{\min},\ldots,r_{\max}\}$ and setting $k_e=N_e-r$,
we re-noise the bridged latent to $\sigma_{k_e}^e$ and let the frozen
teacher run its remaining $r$ steps:
\begin{equation}
    x_{\mathrm{ref}}^e
    =
    \mathcal{G}_e\big(\mathrm{Euler}_{T_e}(\bar{z}_{k_e}^{e},k_e\!\rightarrow\!N_e,c)\big).
    \label{eq:teacher-refine}
\end{equation}
The result $x_{\mathrm{ref}}^e$ is the teacher's correction of the
student's own sample, and $r$ interpolates between the two stages: at
$r=N_e$ it reduces to the off-policy target of Stage 1 (an unconditional
teacher sample), at $r=1$ it degenerates toward the student's own output,
and intermediate $r$ sets how much of the teacher's trajectory overrides
the student's. We use
$(r_{\min},r_{\max})=(\texttt{RMIN},\texttt{RMAX})$. 

To compare this target against what the student itself would produce
under the same remaining budget, and since the two schedules are shifted
differently, we align by noise magnitude rather than timestep index: the
student resumes from its cached state
$j=\max\{i:\sigma_i^S\geq\sigma_{k_e}^e>\sigma_{i+1}^S\}$ and recomputes
only the remaining steps with gradients,
\begin{equation}
\begin{aligned}
    &\tilde{x}_S=\mathcal{G}_S\big(\mathrm{Euler}_S(\mathrm{sg}[z_j^S],\,j\!\rightarrow\!N_S,\,c;\theta)\big),\\[2pt]
    &\mathcal{L}_{\mathrm{OPD}}^{e}=1-\cos\!\big(f(\tilde{x}_S),\,\mathrm{sg}[f(x_{\mathrm{ref}}^e)]\big),
\end{aligned}
\label{eq:opd-loss}
\end{equation}
where $f(\cdot)$ is the CLS embedding of a frozen DINOv2 encoder
\citep{oquab2023dinov2}. The refined image is a fixed target, so
gradients flow only through the rerolled segment. Comparing in a
semantic feature space is what makes cross-model supervision feasible~\cite{kang2024distilling}:
the CLS embedding is invariant to the model-specific pixel statistics
that latent-incompatible models cannot share, so the student learns the
semantic content of the teacher's correction rather than its texture
signature. Appendix shows that the two
entry points into the aligned segment, namely the teacher re-noising the
decoded image and the student resuming from its cached state, are
content-consistent to first order once the student is near
convergence, which is why Stage 1 precedes Stage 2.

\paragraph{Mode sampling and objective.}
Each step samples one mode from a fixed prior $\pi$, with
$\pi(\mathrm{pref})=\lambda$ and $\pi(\mathrm{comp})=1-\lambda$ for
$M=2$; the prompt is then drawn from $\mathcal{P}_e$, through the
gap-aware distribution of Sec.~\ref{sec:gap-aware} when
$e=\mathrm{comp}$. The two stages optimize
\begin{equation}
\begin{aligned}
    &\mathcal{L}^{(1)}=\mathbb{E}_{e\sim\pi}\big[\mathcal{L}_{\mathrm{WS}}^{e}\big],\\[2pt]
    &\mathcal{L}^{(2)}=\mathbb{E}_{e\sim\pi}\big[\mathcal{L}_{\mathrm{OPD}}^{e}+\lambda_{\mathrm{WS}}\mathcal{L}_{\mathrm{WS}}^{e}\big],
\end{aligned}
\label{eq:total}
\end{equation}
where the retained warm-start term regularizes against drift. Each step
updates the shared attention LoRA and the FFN adapter of the sampled mode
only; the backbone and inactive adapters receive no gradient. The
per-step cost is one gradient-free student rollout, an $r$-step teacher
refinement, and a backward pass over the $N_S-j$ tail steps; since a
small $r$ places $\sigma_{k_e}^e$ low on the schedule and hence pushes
$j$ toward $N_S$, the same knob $r$ controls both how on-policy the
target is and the training cost.

\subsection{Capability-Selectable Adapters}
\label{sec:moe}

\begin{figure}[t]
\centering
\includegraphics[width=\columnwidth]{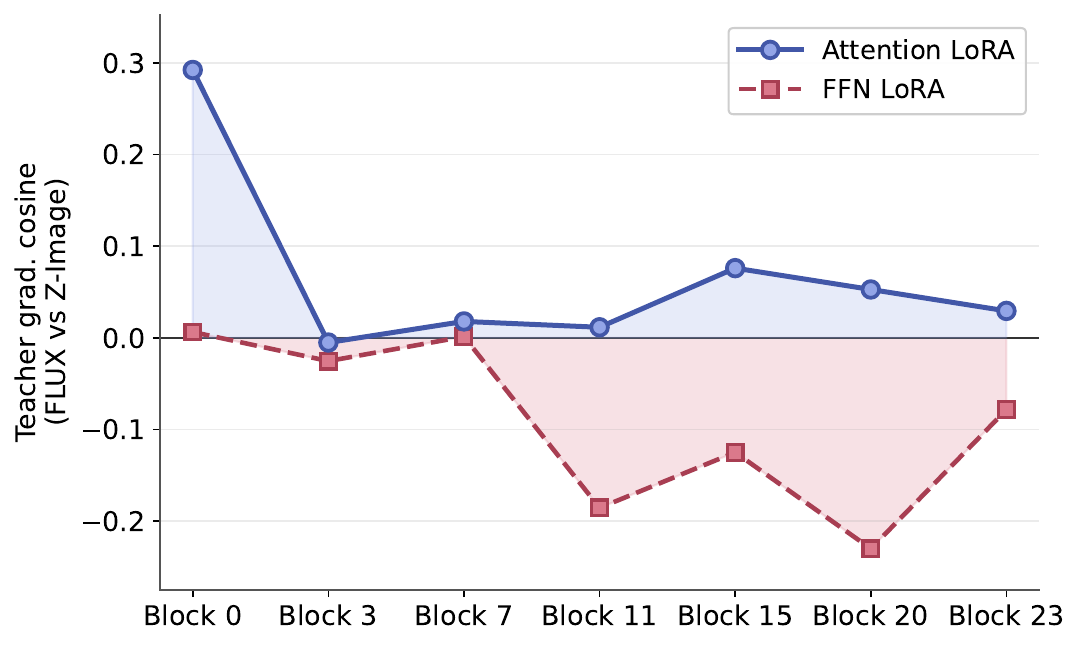}
\vspace{-0.6em} 
\caption{Gradient compatibility diagnostic. Bars report mean cosine
similarity between preference- and composition-mode LoRA gradients over
$N=1{,}000$ paired samples at the warm-start initialization, for four
uniformly sampled blocks of the 24-layer SD3.5-Medium student. Attention
gradients are aligned across modes; FFN gradients are less compatible and
negative in several blocks.}
\label{fig:grad-conflict}
\end{figure}

Multi-teacher OPD poses a sharing--isolation trade-off: a fully shared
adapter mixes preference- and composition-oriented gradients and invites
destructive interference~\cite{yu2020gradient}, while one independent LoRA per capability
duplicates parameters and discards transferable updates. Rather than fix
this boundary by convention, we set it by measurement~\cite{shi2023recon}. Writing
$\mathcal{L}_{\mathrm{pref}}$ and $\mathcal{L}_{\mathrm{comp}}$ for the
mode-conditional losses, instantiated with
Eq.~\eqref{eq:warm-start-loss} so that both gradients share parameters
and loss form, we compute for module type $u$ in block $l$
\begin{equation}
    \kappa_l^u =
    \mathbb{E}\!\left[
    \frac{\langle\nabla_{\theta_l^u}\mathcal{L}_{\mathrm{pref}},\,\nabla_{\theta_l^u}\mathcal{L}_{\mathrm{comp}}\rangle}
    {\|\nabla_{\theta_l^u}\mathcal{L}_{\mathrm{pref}}\|_2\,\|\nabla_{\theta_l^u}\mathcal{L}_{\mathrm{comp}}\|_2}
    \right],
    \label{eq:kappa}
\end{equation}
over paired samples that share $(\sigma,\epsilon)$ across modes and differ
only in prompt source and teacher, so that schedule variance is removed
and any directional difference is attributable to the supervision source. Figure~\ref{fig:grad-conflict} shows a clear asymmetry: attention
gradients remain aligned across modes, while FFN gradients turn negative
in several blocks, consistent with attention carrying shared
text--image routing and feed-forward layers carrying content priors on
which the two objectives disagree~\cite{geva2020transformer}. The measurement directly dictates the design: we share attention
adaptation as the transferable path and isolate FFN adaptation as the
capability-specific one, freezing the backbone and training only LoRA~\cite{hu2022lora}:
parameters:
\begin{equation}
\begin{aligned}
    &W^{\mathrm{attn}}_l \leftarrow W^{\mathrm{attn}}_l + B^{\mathrm{attn}}_l A^{\mathrm{attn}}_l,\\[2pt]
    &W^{\mathrm{ffn}}_l \leftarrow W^{\mathrm{ffn}}_l + B^{\mathrm{ffn}}_{l,e} A^{\mathrm{ffn}}_{l,e}.
\end{aligned}
\label{eq:lora-update}
\end{equation}
The FFN updates
$\phi_e=\{(A^{\mathrm{ffn}}_{l,e},B^{\mathrm{ffn}}_{l,e})\}_{l=1}^{L}$
form the mode-specific adapter. Since only the FFN path is duplicated,
the parameter cost of an additional capability is sublinear in the
number of modes; one frozen backbone carries every consolidated
capability, and switching between them at inference costs an adapter
swap rather than a second model.

\begin{table*}[!t]
\centering
\small
\setlength{\tabcolsep}{3pt}
\caption{Main results on DrawBench using the preference mode of \method{}.
Aesth.: Aesthetic Score; ImgRwd:
ImageReward; HPSv3: Human Preference Score v3; UR: UnifiedReward.
$^\dagger$ denotes a teacher reference; blue shading marks improvement
over SD3.5-Medium.}
\label{tab:main-drawbench}
\renewcommand{\arraystretch}{0.82}
\begin{tabular*}{\textwidth}{@{\extracolsep{\fill}}lccccccc@{}}
\toprule
Model
& Aesth.$\uparrow$
& ImgRwd$\uparrow$
& PickScore$\uparrow$
& HPSv3$\uparrow$
& UR-Alignment$\uparrow$
& UR-Coherence$\uparrow$
& UR-Style$\uparrow$ \\
\midrule
SD3.5-Medium (2.5B)
& 5.383 & 0.922 & 0.866 & 9.341 & 3.057 & 3.435 & 3.011 \\
FLUX.1-dev$^\dagger$ (12B)
& \textbf{5.717} & 0.916 & \textbf{0.878} & \textbf{11.925} & 3.199 & \textbf{3.669} & \textbf{3.261} \\
Z-Image$^\dagger$ (6B)
& 5.261 & 0.830 & 0.856 & 9.076 & 3.183 & 3.506 & 3.040 \\
\midrule
\textbf{\method{}-Preference (2.5B)}
& \scorehl{5.464}
& \textbf{\scorehl{1.168}}
& \scorehl{0.869}
& \scorehl{11.354}
& \textbf{\scorehl{3.206}}
& \scorehl{3.503}
& \scorehl{3.115} \\
\bottomrule
\end{tabular*}
\renewcommand{\arraystretch}{1.0}
\vspace{0.35em}

\footnotesize
\setlength{\tabcolsep}{2pt}
\caption{Main results on GenEval and DPG-Bench using the composition mode of
\method{}. Pos.: position;
2-Obj.: two-object; Attr.: attribute; Ent.: entity; Glob.: global;
Rel.: relation. $^\dagger$ denotes a teacher reference; blue shading
marks improvement over SD3.5-Medium.}
\label{tab:main-composition}
\renewcommand{\arraystretch}{0.82}
\begin{tabular*}{\textwidth}{@{\extracolsep{\fill}}l|ccccccc|cccccc@{}}
\toprule
& \multicolumn{7}{c|}{GenEval}
& \multicolumn{6}{c}{DPG-Bench} \\
\cmidrule(lr){2-8}
\cmidrule(lr){9-14}
Model
& Overall & Count & Pos. & 2-Obj. & Color & Attr. & Single
& Overall & Attr. & Ent. & Glob. & Other & Rel. \\
\midrule
SD3.5-M (2.5B)
& 67.30 & 59.06 & 24.00 & 82.07 & 81.12 & 58.75 & 99.06
& 84.51 & 82.07 & 87.54 & 81.35 & 70.90 & 82.37 \\
FLUX$^\dagger$ (12B)
& 65.20 & \textbf{70.00} & 19.50 & 80.30 & 77.90 & 44.30 & 99.40
& 82.96 & 80.03 & 86.55 & 79.97 & 73.30 & 81.59 \\
Z-Image$^\dagger$ (6B)
& 69.40 & 62.50 & \textbf{32.00} & 86.90 & 83.80 & 52.80 & 98.40
& \textbf{86.08} & \textbf{83.92} & \textbf{88.83} & \textbf{85.50} & \textbf{81.91} & 84.69 \\
\midrule
\textbf{\method{}-Composition (2.5B)}
& \textbf{\scorehl{73.30}}
& \scorehl{59.90}
& \scorehl{31.20}
& \textbf{\scorehl{97.00}}
& \textbf{\scorehl{84.60}}
& \textbf{\scorehl{67.00}}
& \textbf{\scorehl{100.00}}
& \scorehl{85.80}
& \scorehl{83.14}
& \scorehl{88.60}
& \scorehl{83.25}
& \scorehl{78.62}
& \textbf{\scorehl{85.20}} \\
\bottomrule
\end{tabular*}
\renewcommand{\arraystretch}{1.0}
\end{table*}

\subsection{Gap-Aware Adaptive Sampling}
\label{sec:gap-aware}

The compositional mode is itself non-uniform: subskills such as single
object or color saturate early while position and attribute binding
remain bottlenecks, a pattern visible in the teacher scores of
Table~\ref{tab:main-composition}. Uniform sampling therefore spends equal
budget on saturated and unsolved categories, yet sampling by raw
difficulty is brittle, since a low teacher score need not imply a useful
signal. We instead allocate budget by the remaining
\textbf{teacher--student gap} rather
than the absolute difficulty~\cite{zhou2021curriculum}.

For each compositional category $a\in\mathcal{A}$ we hold out $k$ probe
prompts, disjoint from both the training pool and the evaluation set, and
cache an offline teacher score $s_a^T$. Every $K$ steps the student is
scored on the same probes to obtain $s_a^S$, and the positive residual is
smoothed and normalized into a category distribution,
\begin{equation}
\begin{aligned}
    &g_a\leftarrow\beta g_a+(1-\beta)\max(s_a^T-s_a^S,0),\\[2pt]
    &p(a)=\frac{\exp(g_a/\tau)}{\sum_{a'}\exp(g_{a'}/\tau)}.
\end{aligned}
\label{eq:gap}
\end{equation}
Clipping at zero distinguishes this from a difficulty-based curriculum:
a matched category carries no exploitable margin, so its weight decays
regardless of its absolute score, and the allocation anneals back toward
uniform without an external schedule. The temperature $\tau$ sets how
greedily the largest gap is favored, and the momentum $\beta$ smooths
the $k$-probe estimates against measurement noise. Each compositional
batch samples $a\sim p(a)$ and draws prompts from the corresponding
bucket; preference-mode prompts are sampled independently from their
fixed distribution.

\begin{table*}[!tp]
\centering
\small
\setlength{\tabcolsep}{3pt}
\caption{DrawBench ablation of the three components, evaluated with the
preference adapter. \emph{w/o CSA} replaces capability-specific FFN
adapters with one shared adapter, \emph{w/o gap-aware} samples
composition categories uniformly, and \emph{w/o warm start} runs OPD
directly from the base student, in each case keeping everything else
fixed. }
\label{tab:ablation-drawbench}
\begin{tabular*}{\textwidth}{@{\extracolsep{\fill}}lccccccc@{}}
\toprule
Model
& Aesth.$\uparrow$
& ImgRwd$\uparrow$
& PickScore$\uparrow$
& HPSv3$\uparrow$
& UR-Alignment$\uparrow$
& UR-Coherence$\uparrow$
& UR-Style$\uparrow$ \\
\midrule
Full & \scorehl{5.464} & \textbf{\scorehl{1.168}} & \scorehl{0.869} & \textbf{\scorehl{11.354}} & \textbf{\scorehl{3.206}} & \scorehl{3.503} & \textbf{\scorehl{3.115}} \\
w/o gap-aware & 5.546 & 0.966 & \textbf{0.873} & 11.214 & 3.164 & \textbf{3.530} & 3.113 \\
w/o CSA & 5.479 & 0.804 & 0.858 & 10.117 & 3.111 & 3.480 & 3.074 \\
w/o warm start & \textbf{5.635} & 0.860 & 0.857 & 9.828 & 3.103 & 3.478 & 3.106 \\
\bottomrule
\end{tabular*}
\end{table*}

\begin{table*}[!tp]
\centering
\footnotesize
\setlength{\tabcolsep}{2pt}
\caption{GenEval and DPG-Bench ablation of adapter design, gap-aware
sampling, and warm start. Pos.: position; 2-Obj.: two-object; Attr.:
attribute; Ent.: entity; Glob.: global; Rel.: relation. Blue shading
marks the full configuration.}
\label{tab:ablation-composition}
\begin{tabular*}{\textwidth}{@{\extracolsep{\fill}}l|ccccccc|cccccc@{}}
\toprule
\multirow{2}{*}{Model}
& \multicolumn{7}{c|}{GenEval}
& \multicolumn{6}{c}{DPG-Bench} \\
\cmidrule(lr){2-8}
\cmidrule(lr){9-14}
& Overall & Count & Pos. & 2-Obj. & Color & Attr. & Single
& Overall & Attr. & Ent. & Glob. & Other & Rel. \\
\midrule
Full
& \textbf{\scorehl{73.3}} & \textbf{\scorehl{59.9}} & \textbf{\scorehl{31.2}} & \textbf{\scorehl{97.0}} & \scorehl{84.6} & \textbf{\scorehl{67.0}} & \textbf{\scorehl{100.0}}
& \textbf{\scorehl{85.80}} & \textbf{\scorehl{83.14}} & \textbf{\scorehl{88.60}} & \textbf{\scorehl{83.25}} & \textbf{\scorehl{78.62}} & \textbf{\scorehl{85.20}} \\
w/o gap-aware
& 68.5 & 53.8 & 30.8 & 88.4 & 84.6 & 54.8 & 99.1
& 83.76 & 81.39 & 86.71 & 80.23 & 78.15 & 81.65 \\
w/o CSA
& 62.3 & 39.7 & 17.2 & 82.1 & \textbf{87.2} & 48.5 & 99.1
& 83.59 & 81.40 & 86.58 & 78.71 & 72.35 & 82.19 \\
w/o warm start
& 49.2 & 44.1 & 20.5 & 63.6 & 58.0 & 25.3 & 84.1
& 81.00 & 78.01 & 84.68 & 81.22 & 70.04 & 78.05 \\
\bottomrule
\end{tabular*}
\end{table*}

\begin{figure*}[!t]
\centering
\includegraphics[width=0.88\textwidth]{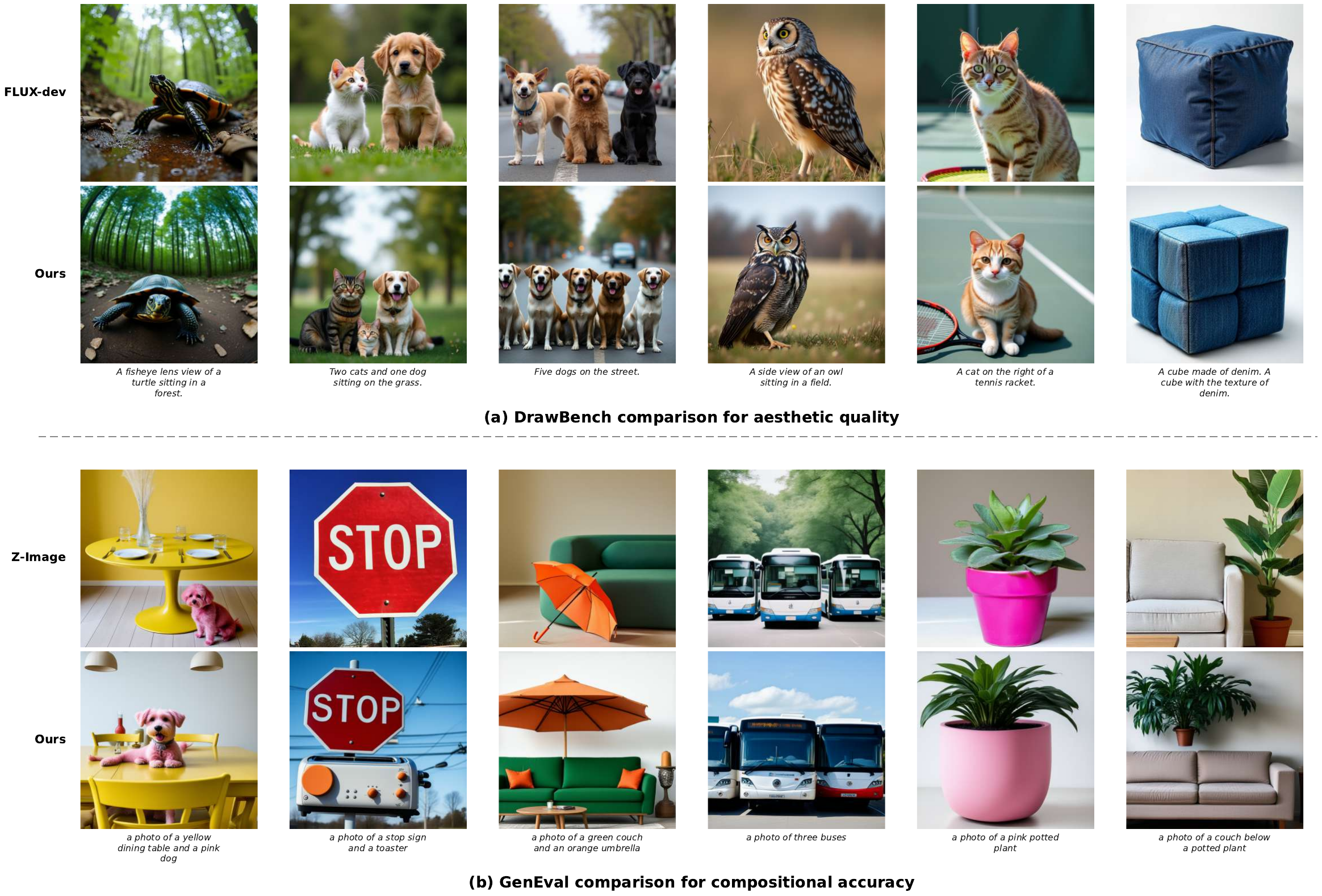}
\caption{Qualitative comparison on representative prompts. \textbf{(a) DrawBench
comparison for aesthetic quality}: FLUX produces visually appealing images but
can miss prompt constraints, such as the dog count in the third column and the
cat--tennis-racket spatial relation in the fifth column. \textbf{(b) GenEval comparison
for compositional accuracy}: Z-Image follows many compositional prompts but can
omit required objects, such as the toaster in the second column, or produce less
pleasing appearances, such as the over-saturated pink object in the fifth
column. \method{} combines the complementary teacher strengths, preserving
stronger visual quality while improving object presence, counts, colors, and
spatial relations.}
\label{fig:qualitative}
\end{figure*}

\begin{table*}[!tp]
\centering
\begin{minipage}[t]{0.47\textwidth}
\centering
\scriptsize
\setlength{\tabcolsep}{2.8pt}
\caption{Ablation of the teacher refinement range $[r_{\min},r_{\max}]$
in Eq.~\eqref{eq:teacher-refine} on the 20-step teacher grid, with all
other components fixed. 
}
\label{tab:krange}
\resizebox{\linewidth}{!}{%
\begin{tabular}{lccccc}
\toprule
Refinement range & GenEval & DPG & ImgRwd & HPSv3 & UR-Avg \\
\midrule
High noise, $[15,20]$ & \textbf{\scorehl{73.3}} & \textbf{\scorehl{85.80}} & \textbf{\scorehl{1.168}} & \textbf{\scorehl{11.354}} & \scorehl{3.275} \\
Middle noise, $[10,15]$ & 70.5 & 82.74 & 0.990 & 11.218 & 3.304 \\
Low noise, $[0,10]$ & 67.2 & 81.58 & 1.001 & 11.113 & \textbf{3.323} \\
\bottomrule
\end{tabular}}
\end{minipage}
\hfill
\begin{minipage}[t]{0.49\textwidth}
\centering
\scriptsize
\setlength{\tabcolsep}{2.8pt}
\caption{Ablation of the perceptual space $f(\cdot)$ in
Eq.~\eqref{eq:opd-loss}, replacing the frozen DINOv2 encoder while
keeping the teacher pool, adapters, sampling, and refinement range
fixed.  }
\label{tab:feature-space}
\resizebox{\linewidth}{!}{%
\begin{tabular}{lccccc}
\toprule
Feature & GenEval & DPG & ImgRwd & HPSv3 & UR-Avg \\
\midrule
Student latent MSE & \multicolumn{5}{c}{\ding{55}\ Collapse} \\
ConvNeXt & 66.2 & 81.54 & 0.936 & 11.192 & 3.266 \\
SigLIP & 65.8 & 84.62 & 0.850 & 11.338 & \textbf{3.291} \\
DINOv2 & \textbf{\scorehl{73.3}} & \textbf{\scorehl{85.80}} & \textbf{\scorehl{1.168}} & \textbf{\scorehl{11.354}} & \scorehl{3.275} \\
\bottomrule
\end{tabular}}
\end{minipage}
\end{table*}

\section{Experiments}
\label{sec:exp}

\subsection{Experimental Setup}

\paragraph{Models and training.}
SD3.5-Medium~\cite{esser2024sd3} is the 2.5B student, distilled from
FLUX.1-dev~\cite{flux2024} for the preference mode and
Z-Image~\cite{cai2025z} for the composition mode, with prompts drawn
respectively from Pick-a-Pic~\cite{kirstain2023pickscore} and the
GenEval-style split of Flow-GRPO~\cite{liu2026flow}. Warm start runs 500
steps, followed by 800 OPD steps with a 20-step student Euler sampler,
DINOv2 CLS supervision, and refinement depths $r\in[15,20]$ on the
20-step teacher grid (Eq.~\eqref{eq:teacher-refine}). Gap-aware sampling
covers five composition categories (two-object, counting, colors,
position, color attribution) with a held-out probe set of $k=32$ prompts
each and cached teacher scores; every $K=50$ steps we score the probes,
update the EMA gaps with $\beta=0.8$, and refresh $p(a)$ with
$\tau=0.1$.

\paragraph{Evaluation protocol.}
All models are evaluated at $512\times512$, with teachers using their
official sampling settings and the student and \method{} the
SD3.5-Medium protocol; each benchmark is evaluated with its matched
adapter, fixed across all methods and checkpoints.
DrawBench~\cite{saharia2022drawbench} (preference adapter, 999 prompts,
one sample each)

reports Aesthetic score~\cite{schuhmann2022laion},
ImageReward~\cite{xu2023imagereward},
PickScore~\cite{kirstain2023pickscore}, HPSv3~\cite{ma2025hpsv3}, and
UnifiedReward~\cite{wang2025unified}. GenEval~\cite{ghosh2023geneval}
(553 prompts) and DPG-Bench~\cite{hu2024ella} (1065 prompts) use the
composition adapter with four samples per prompt, reporting overall
scores alongside per-category breakdowns. \textbf{Full sampler, guidance, and
schedule settings are in the supplement.}

\subsection{Main Results}
Tables~\ref{tab:main-drawbench} and~\ref{tab:main-composition} compare
\method{} against the original student and the two teacher references.
A single 2.5B student improves on both axes at once, surpassing both
larger teachers on GenEval and its FLUX.1-dev teacher on ImageReward
and UR-Alignment; exceeding a supervising teacher is a signature of
on-policy training, since fitting a teacher-generated corpus is bounded
by the teacher's own sample distribution.

\paragraph{Preference mode.}
On DrawBench, every preference metric improves over
SD3.5-Medium: ImageReward rises from 0.922 to 1.168 and HPSv3 from
9.341 to 11.354, with the three UnifiedReward sub-scores gaining 0.149,
0.068, and 0.104. PickScore is essentially unchanged (0.869 vs.\
0.866): the metric closest to the student's original training signal
neither gains nor regresses. Against the 12B teacher itself, the 2.5B
student surpasses FLUX.1-dev on ImageReward and UR-Alignment and
recovers about 78\% of the HPSv3 margin between student and teacher.

\paragraph{Composition mode.}
GenEval overall accuracy rises from 67.30 to 73.30, exceeding the
Z-Image teacher (69.40) by 3.90 points and FLUX.1-dev (65.20) by 8.10.
The gains concentrate on structured categories: two-object composition
improves by 14.93 points, attribute binding by 8.25, and position by
7.20, against 3.48 for color and 0.84 for counting, and
Sec.~\ref{sec:ablation} confirms that the two-object, attribution, and
counting gains depend on gap-aware allocation. Attribute binding is
also the category where the student ends clearly above both teachers
(67.00 vs.\ 52.80 and 44.30), showing that the student is not capped by
its teachers' per-category scores. On DPG-Bench the student
improves from 84.51 to 85.80, coming within 0.28 points of the Z-Image
teacher while achieving the best relation score (85.20), in line with the same structured-category pattern.
Figure~\ref{fig:qualitative} shows qualitative examples matching these
trends.

\subsection{Ablation Studies}
\label{sec:ablation}

\paragraph{Architecture and sampling.}
Tables~\ref{tab:ablation-drawbench} and~\ref{tab:ablation-composition}
ablate the two design choices: \emph{w/o CSA} replaces
capability-specific FFN adapters with one shared adapter, and \emph{w/o
gap-aware} samples GenEval categories uniformly, in both cases keeping
everything else fixed. Removing adapter selection is the most damaging,
costing 11.0 GenEval points, 1.24 HPSv3, 0.364 ImageReward, and 2.21 on
DPG-Bench, which indicates that a single shared adapter indeed averages
away the conflicting FFN updates. Removing gap-aware sampling mainly
hurts the categories with the largest gaps, two-object (88.4 vs.\
97.0), color attribution (54.8 vs.\ 67.0), and counting (53.8 vs.\
59.9), showing that adaptive allocation is what lets the student close
its remaining compositional deficits rather than a uniform polish.

\begin{figure}[t]
\centering
\includegraphics[width=0.82\columnwidth]{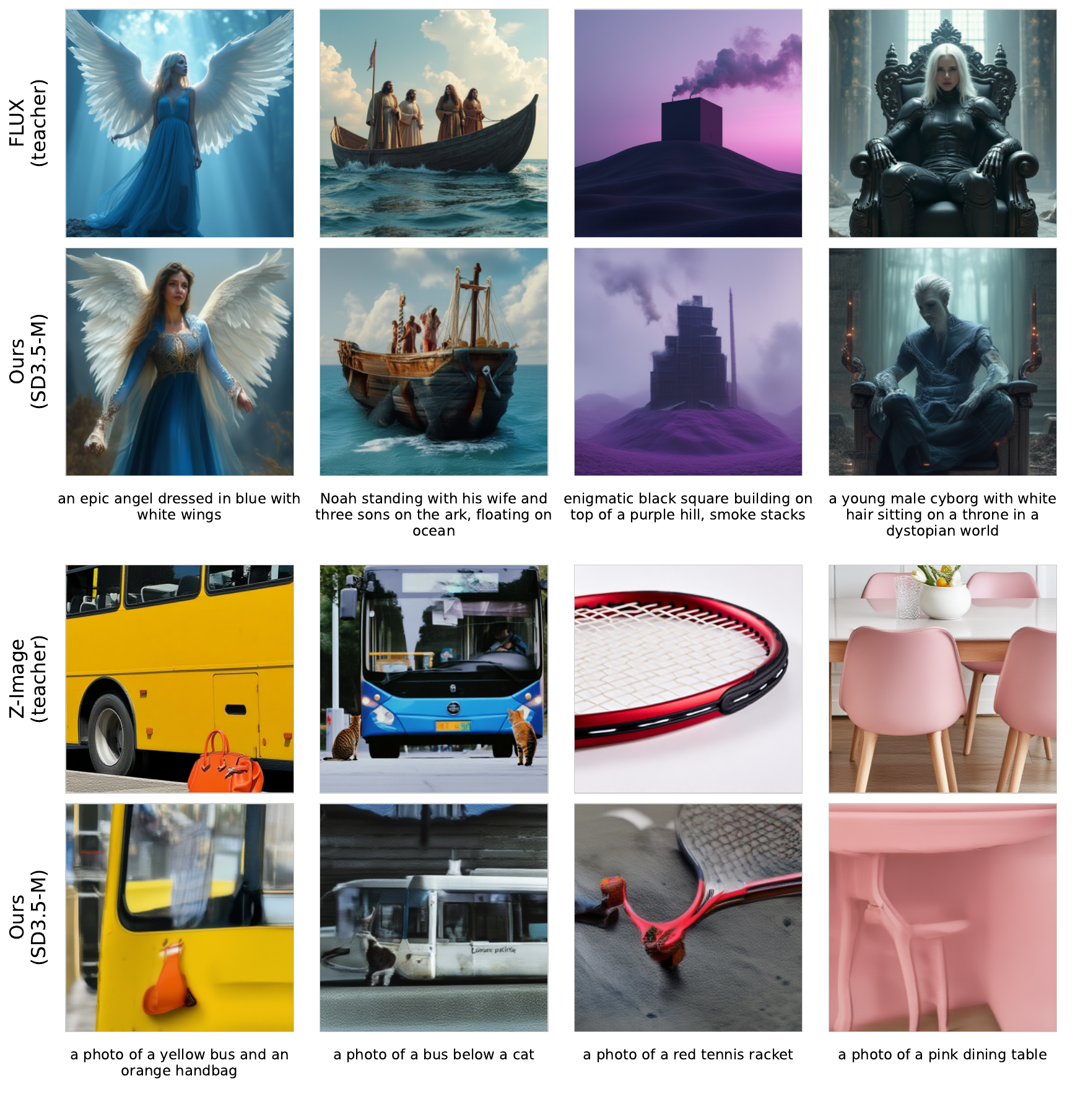}
\caption{Qualitative effect of the warm-start stage. }
\label{fig:supp-sft-qualitative}
\end{figure}

\paragraph{Warm start.}
The \emph{w/o warm start} variant removes Stage 1 and starts OPD
directly from the base student. It degrades every metric, by 24.1
GenEval points, 4.80 on DPG-Bench, 0.308 ImageReward, and 1.526 HPSv3,
indicating that teacher refinement only becomes a reliable signal once
the student distribution has moved toward the teachers'. The warm-start
stage alone fails in the opposite direction: trained only on
teacher-generated images, the student produces over-smoothed outputs
and misses fine detail (Figure~\ref{fig:supp-sft-qualitative}). We
therefore treat warm start as an initialization that stabilizes
subsequent OPD.

\paragraph{Refinement noise.}
Table~\ref{tab:krange} varies the teacher refinement range. Preference
metrics are largely insensitive (HPSv3 within 0.24, UR-Avg within
0.05), whereas GenEval rises from 67.2 to 73.3 as the range moves from
$[0,10]$ to $[15,20]$. Compositional prompts thus require the teacher
to intervene at larger noise levels, where layout, object count, and
spatial relations can still be reorganized, while appearance quality is
already served by a low-noise local polish.

\paragraph{Perceptual representation.}
Replacing DINOv2 with ConvNeXt~\cite{liu2022convnet} or
SigLIP~\cite{zhai2023sigmoid}, all else fixed, keeps preference quality
competitive but costs 7.1 and 7.5 GenEval points respectively
(Table~\ref{tab:feature-space}), suggesting that the self-supervised
ViT representation carries a stronger structural signal for
object-centric distillation. A direct MSE loss in student latent
coordinates collapses outright, confirming that OPD needs a stable
common feature space rather than raw coordinate matching.

\section{Conclusion}
\label{sec:conclusion}
We presented \method{}, a framework for on-policy distillation from
heterogeneous teachers whose latent spaces and noise schedules are
incompatible with those of the student. Its
pixel bridge turns such teachers into sources of on-policy supervision,
its adapters set the sharing--isolation boundary by gradient
measurement, and its gap-aware sampling anneals
itself as teacher--student residuals close; one frozen backbone thus
carries every capability, and selection costs an adapter swap. Two of
these ideas travel beyond our setting: the refinement depth
interpolates continuously between off-policy imitation and on-policy
correction, and measured sharing boundaries apply to any multi-teacher
adaptation. A 2.5B student raises GenEval from 67.3 to 73.3, above both
larger teachers, and DrawBench HPSv3 from 9.34 to 11.35; surpassing the
supervising teachers at all is evidence unique to on-policy training,
since imitating teacher samples is bounded by the teachers themselves.

\bibliography{references}

\end{document}